\documentclass[journal]{IEEEtran}

\usepackage{microtype}
\usepackage{graphicx}
\usepackage{subcaption}

\usepackage[utf8]{inputenc} 
\usepackage[T1]{fontenc}    
\usepackage{hyperref}       
\usepackage{url}            
\usepackage{amsfonts}       
\usepackage{amsmath}
\usepackage{nicefrac}       
\usepackage{microtype}      
\usepackage{xcolor}         
\usepackage{listings}

\usepackage{multirow}
\usepackage{bbm}
\usepackage[font={small}]{caption}
\usepackage{wrapfig}
\usepackage{enumitem}

\usepackage{booktabs}

\usepackage{hyperref}

\usepackage{amsmath}
\usepackage{amssymb}
\usepackage{mathtools}
\usepackage{amsthm}

\usepackage[capitalize,noabbrev]{cleveref}

\theoremstyle{plain}

\theoremstyle{definition}

\theoremstyle{remark}

\usepackage[textsize=tiny]{todonotes}

\begin{document}



\title{Pretraining Graph Neural Networks for few-shot Analog Circuit Modeling and Design}

\author{%
  Kourosh Hakhamaneshi \textsuperscript{$\ast$}\textsuperscript{$\dagger$}, 
  Marcel Nassar \textsuperscript{$\dagger$}, 
  Mariano Phielipp \textsuperscript{$\dagger$}, \\
  Pieter Abbeel \textsuperscript{$\ast$},
  Vladimir Stojanovic \textsuperscript{$\ast$} \\
  $\ast$ University of California, Berkeley \\
  $\dagger$ Intel AI Labs\\
}
\maketitle





%
\IEEEpeerreviewmaketitle

\begin{abstract}
Being able to predict the performance of circuits without running expensive simulations is a desired capability that can catalyze automated design.
In this paper, we present a supervised pretraining approach to learn circuit representations that can be adapted to
new circuit topologies or unseen prediction tasks.
We hypothesize that if we train a neural network (NN) that can predict the output DC voltages of a wide range of circuit instances it will be forced to learn generalizable knowledge about the role of each circuit element and how they interact with each other. 
The dataset for this supervised learning objective can be easily collected at scale since the required DC simulation to get ground truth labels is relatively cheap.
This representation would then be helpful for few-shot generalization to unseen circuit metrics that require more time consuming simulations for obtaining the ground-truth labels. 
To cope with the variable topological structure of different circuits we describe each circuit as a graph and use graph neural networks (GNNs) to learn node embeddings.
We show that pretraining GNNs on prediction of output node voltages can encourage learning representations that can be adapted to new unseen topologies or prediction of new circuit level properties with up to 10x more sample efficiency compared to a randomly initialized model.
We further show that we can improve sample efficiency of prior SoTA model-based optimization methods by 2x (almost as good as using an oracle model) via fintuning pretrained GNNs as the feature extractor of the learned models.
\end{abstract}


\section{Introduction}

\IEEEPARstart{T}{he}
semiconductor industry largely owes its rapid progress in the last century to the scaling of silicon process technology, which has consistently allowed designers to fit more processing power into the same area footprint and make today's digital processing capabilities a reality. However, such highly specialized designs often come at a high re-design cost with every new integrated circuits (ICs) technology. This is often due to the change of low level transistor behaviors and more stringent design rules imposed by technology scaling.  
Recent research efforts have focused on shortening the production cycle by leveraging the advances made in machine learning in several aspects of the manual design process, 
including high-level synthesis \cite{makrani2019xppe}, logic synthesis \cite{haaswijk2018deep, hosny2020drills}, placement and routing \cite{mirhoseini2021graph, lu2019gan, xie2018routenet}, analog mixed signal layout exploration \cite{wang2020gcn, settaluri2020autockt, hakhamaneshi2019bagnet, li2020customized}, manufacturing \cite{jiang2020efficient, ye2019lithogan}, and others \footnote{For a comprehensive survey on using machine learning in electronic design automation see \cite{huang2021machine}.}.


The main obstacle for large-scale adoption of ML-based methods to circuit design is the difficulty in creating large datasets partly due to expensive simulation run-times. For example the simulation of one instance of a digital to analog converter (DAC) in ~\cite{hassanpourghadi2021circuit} could take $\sim 30$ minutes and this time can become as much as a few days for more complicated circuits. 
To mitigate this issue, prior work has explored applying adaptive sampling methods such as reinforcement learning \cite{settaluri2020autockt}, bayesian optimization \cite{hakhamaneshi2021jumbo, lyu2018batch}, and evolutionary algorithms \cite{hakhamaneshi2019bagnet} for optimization of circuits.
However, the models or policies that are learned via these methods are highly tailored to the environment and objective of the optimization and cannot be re-used for new specifications, topologies, or more advanced technologies.
Ideally we want to be able to re-use the representation learned from one problem to solve another problem faster, similar to how humans re-use their prior experience to design new circuits.  


In this work, we focus on modeling analog circuits and their generalization to unseen circuit topologies.
Such a generalizable model can help boost the performance of model-based optimization methods in terms of sample efficiency and performance. 
Our main research question is \textit{How can we learn circuit representations that can be re-used for modeling unseen topologies or predicting new circuit-level labels via finetuning?} 
We hypothesize that if we train a neural network that can predict all output voltage nodes in a circuit, some domain knowledge will be encoded in the representation that can also be re-used for downstream scenarios.
Predicting node voltages in a circuit requires an in-depth understanding of the static behavior of each element and their interaction with one another.
At the same time, the simulation required for computing the ground-truth output DC voltages is a lot cheaper than getting labels for other performance metrics (e.g. settling time). 
Also knowing node DC voltage values, a.k.a the bias of the circuit, can simplify deriving its behavior and performance. To be able to handle the variable input structure of the model we represent each circuit instance with a graph and use graph neural networks (GNN) as the backbone of the model. 
This allows us to improve generalization by using the structural information in circuits as an input feature for prediction.


We first test this hypothesis with a set of controlled experiments on simple circuits that include only resistors and voltage sources, known as resistor ladders. 
This class of circuits despite being simple from an analytical point of view, have a special property that any change to any part of the circuit (e.g. changing voltage source or resistor values or adding or removing extra elements) can translate into a very different output response on node voltages.
In other words, no node is isolated from another one, making their interaction really challenging for neural networks to model. 
We show that a GNN backbone pretrained with the voltage prediction task on a subset of this circuit family can be fine-tuned to generalize its prediction to new unseen circuits with different topologies. 
We also illustrate how the node representation can be used for new graph property prediction tasks such as prediction of the output resistance of the circuit.
Furthermore, we provide more empirical evidence on a more complicated real-world transistor-based circuit and contrast the sample efficiency of our pretrained models to training from scratch on new unseen tasks. 
We then show that such pretrained GNN model can be reused as the feature extractor in an optimization loop that automates the design of such circuit, and improve the optimization sample efficiency by at least 2x (almost on par with the oracle model) compared to previous state of the art that uses simple fully connected architectures trained from scratch. 

We summarize our contribution as follows:
(1) We present an enhanced representation for describing circuits as graphs that is naturally more tailored to transistor-based topologies compared to prior work on using GNNs on analog circuits \cite{ren2020paragraph, ma2019high}. (2) We show that node voltage prediction can serve as an effective pretraining task for learning features transferable to new topologies in both zero and few-shot manner (3) We further show that such a representation can also be re-used for graph prediction tasks via fine-tuning on small datasets and improve sample efficiency of model-based optimization algorithms (4) We present the first open-sourced graph dataset and benchmark for circuit voltage prediction that can help standardize progress of this research direction.

\section{Related Work}
\label{sec:rel-work}

\textbf{Modeling circuits using deep neural networks}: Functional modeling of analog circuits using deep neural networks has been revisited in recent years due to the success of deep learning in modeling black-box functions.
To this end, a conventional modeling approach is to model the entire circuit end-to-end with multi-layer perceptrons \cite{wolfe2003extraction, li2019artificial}. 
However, this choice of architecture requires large training dataset for building accurate models which can be problematic when simulation is time-consuming.
One way to address this issue has been to impose domain specific inductive biases on model architectures. 
For example,  ~\cite{hassanpourghadi2021circuit} proposes breaking the circuit into sub-circuits and modeling each with a separate NN in an auto-regressive fashion. 
However, it requires ad-hoc human input on the definition of sub-circuits. Also the sub-circuits themselves are modeled with MLPs and do not leverage structural information.
In this work we utilize GNNs, which generalize of the idea of imposing structural inductive bias on the architecture.

\textbf{GNNs in chip design}: In recent years, the idea of using GNNs to automate the design process of systems on chip have been visited in several prior works from both an optimization and modeling perspective. 
Most notably ~\cite{mirhoseini2021graph} studies the problem of chip placement and utilizes GNNs as the backbone of the feature embeddings of an RL policy and value function that is learned in an end-to-end fashion using custom reward functions. 
The reason behind using GNNs as feature extractors is that many chip placement strategies are motivated by the \textit{global} structure of the system and GNNs can capture the structural dependencies effectively.
~\cite{ren2020paragraph}, on the other hand, uses GNNs for circuit parasitic prediction based on circuit's graph which only needs \textit{local} information.
~\cite{liu2021parasitic, zhang2019circuit} employ GNNs as the backbone of the predictive models for optimizing analog circuit metrics. 
However, the benefits of using GNNs in these works are limited to a single topology or a single optimization objective. In our work, we study pretraining GNNs to see whether we can reliably transfer domain knowledge to new topologies or prediction tasks instead of learning from scratch.




\section{Learning representations for analog circuits}

\begin{figure*}[th]
    \centering
    \begin{subfigure}{0.26\textwidth}
        { \includegraphics[width=\textwidth]{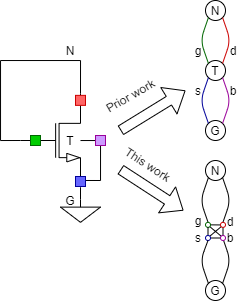}\par\caption{}\label{fig:graph-rep-why}}
    \end{subfigure}
    \begin{subfigure}{0.50\textwidth}
        {\includegraphics[width=\textwidth]{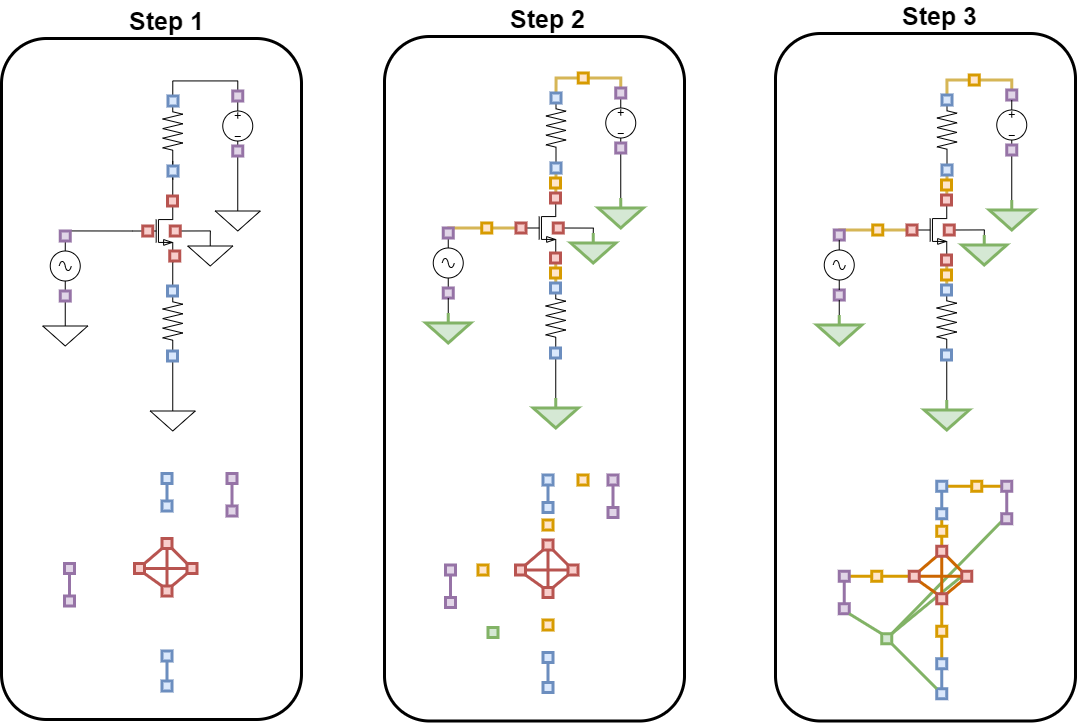}\par\caption{}\label{fig:graph-rep-example}}
    \end{subfigure}
    \caption{
    (a) Example of why it is important to explicitly model device terminals as nodes to disambiguate certain topological configurations. (b) Steps of converting a circuit schematic to a graph representation. (1) Convert all devices to their corresponding complete sub-graphs (2) Convert all circuit nets to (yellow) nodes in the graph (3) Make edge connections according to the topology of the circuit.
    }
\end{figure*}

We will first present the language for describing analog circuits as graphs in section~\ref{sec:graph-rep}.
In section \ref{sec:pretraining-task} and \ref{sec:node-to-graph} we describe how we can use a supervised node property prediction task (i.e. predicting voltages of each node) as a pretraining objective to learn representations that can be fine-tuned for predicting voltages on unseen topologies or for new graph property prediction tasks (e.g. predicting the output resistance of the circuit). We finally conclude by explaining our evaluation protocol in section \ref{sec:eval}.




\subsection{New graph representation}
\label{sec:graph-rep}
Despite the recent surge of interest in utilizing GNNs for predictive models on circuits \cite{ren2020paragraph, ma2019high}, there is still no unified descriptive language for representing transistor-based circuits as graphs that can uniquely represent all possible topological configurations. 
For example, \cite{ren2020paragraph} proposes to map each device and each net (i.e. wire connection) in the circuit to a graph node and uses edge types to distinguish the device terminal types (i.e. source, drain, gate, body). 
This representation becomes ambiguous in cases where a device (e.g. transistor) has the same net connection between two or more terminals. 
Figure \ref{fig:graph-rep-why} illustrates such an example. 
Since the drain (red) and gate (green) terminals of the transistor are both connected to net N, two edges with different types should connect the device T to its corresponding net N. 
The source (blue) and body (purple) terminals also have the same ambiguity. 

To disambiguate this representation, we propose a set of steps that explicitly represent the device terminals as individual nodes.
This representation is closely linked to the widely used textual description (i.e. netlist) of circuits used for simulator tools such as Spice \cite{quarles1994spice}. 
In this representation, each device is a \textit{complete} sub-graph with its terminals as individual nodes. 
The features of each node indicate the terminal type and device parameters and therefore, all edges become feature-less. 
Figure \ref{fig:graph-rep-example} illustrates an example of how a simple amplifier circuit gets mapped to its graph representation, step by step following our approach:

\begin{enumerate}[labelwidth=!, labelindent=0pt, nosep]
    \small
    \item Convert the terminals of each device to a complete sub-graph with the same number of nodes as the number of terminals. 
    \item Convert each circuit net to a node in the graph.
    \item Connect all the nets to their corresponding device terminals.
\end{enumerate}

The resulting representation is a graph $\mathcal{G = (V, E)}$ comprised of a set of nodes $\mathcal{V}$ and a set of undirected edges $\mathcal{E}$. Each node is associated with a feature vector which describes the type of the node as well as the device parameters that the node belongs to (e.g. the value of a resistor). There are no features associated with edges, and they only preserve the connectivity information of the elements of the circuit. 


\begin{figure*}[th]
    \centering
    \includegraphics[width=0.8\textwidth]{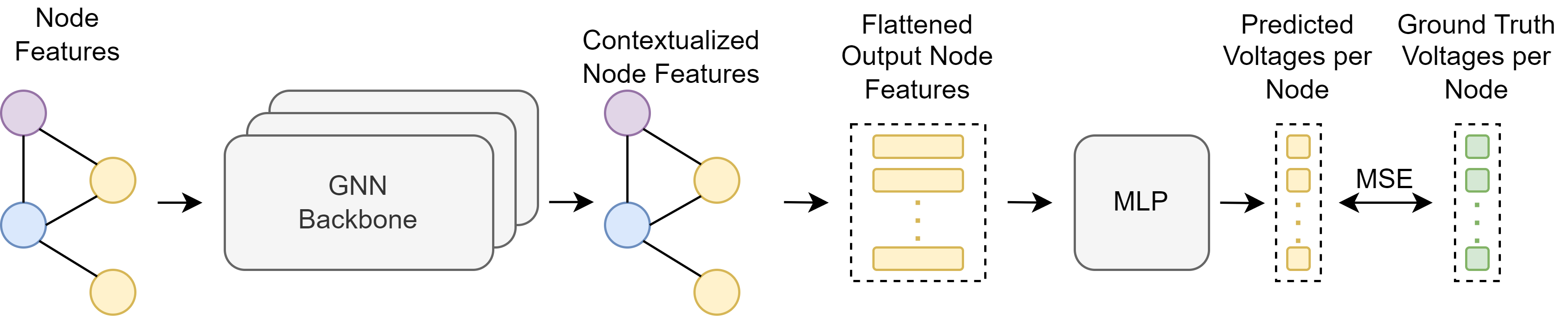}
    \caption{Pretraining architecture: After the GNN backbone, the batch of output node features (yellow nodes) are fed into an MLP to predict the output voltages of each node. The same MLP is applied to all the node features. The gradient of the mean square error between ground truth and prediction is then backpropagated to update the parameters.}
    \label{fig:pretraining}
\end{figure*}

\begin{figure*}[th]
    \centering
    \includegraphics[width=0.8\textwidth]{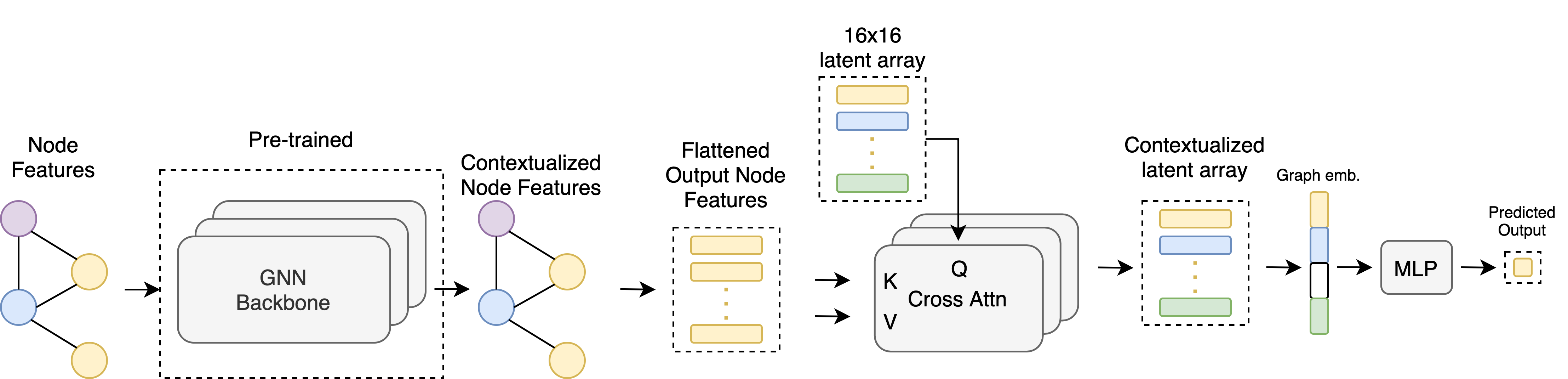}
    \caption{Node to graph prediction pipeline: After the GNN backbone, the node features are translated to a single vector representing the graph embedding. The graph embedding is then fed to an MLP to predict the output. The node to graph embedding is 3 cross attention layers between a learned embedding and the node features (similar to ~\cite{jaegle2021perceiver})}
    \label{fig:finetuning}
\end{figure*}

\subsection{Voltage prediction as the pretraining task}
\label{sec:pretraining-task}

One of the major steps for analyzing and synthesizing circuits for humans is to compute the bias first. 
This entails computing the DC voltage of various important nodes in the circuit that sets the working condition of all the devices. 
One can think of computing voltages as deriving the operating point and linearizing a nonlinear system which is a pre-requisite step for computing other system metrics. 

Inspired by this causal abstraction, we hypothesise that if a neural network is trained to predict the DC voltages of circuit nodes in various contexts it should be able to learn critical system-specific knowledge that can be re-used for other predictions, either in zero-shot or few-shot manner via fine-tuning. 
In other words, the pre-trained backbone on the voltage prediction task should provide a better initialization for unseen tasks and topologies than starting from a random initialization. 
Moreover, getting DC operating points of circuits through simulation is relatively in-expensive compared to simulating other circuit properties.
Therefore, in principle, the pre-training dataset can be collected at scale through simulation in a reasonable time-frame. 

Figure \ref{fig:pretraining} illustrates the pretraining architecture. We cascade a GNN backbone with an MLP with shared parameters across nodes which predicts the output voltage based on the node contextualized representations.
The pretraining objective is minimizing the mean square error (MSE) loss between the predicted output and the ground truth for the output nodes. 
There is no supervision signal on the representation of other non-output nodes. After pretraining, we can use the output node features from GNN backbone for prediction of new tasks or new topologies.

\subsection{From node embedding to graph property prediction}
\label{sec:node-to-graph}

To perform a graph property prediction task, we need to combine the node embeddings into a single graph embedding.
A naive way is to aggregate all the node embeddings via pooling operations (e.g. mean pooling) and then feed it to an MLP that performs the graph level prediction. 
However, this approach can result in loss of information during aggregation and decreased capacity of the neural network. 

To mitigate this problem, we propose a learnable attention-based aggregation layer that scales linearly with the number of output nodes. 
By learning the aggregation module, the model can choose the optimal way of aggregating the embeddings into a graph representation based on the dataset. 

Figure \ref{fig:finetuning} demonstrates this architecture for fine-tuning on a graph property prediction task.
Once we have the contextualized node embeddings at the output of the GNN, we flatten them as a \textit{set} of feature vectors and pass them through a cross attention network.
In this attention module we compute the cross attention between a set of learned latent vectors and the output node features to embed them into a single graph representation.
The latent vectors are represented as a 16x16 array that is learned during training to minimize the output MSE loss.
The contextualized 16x16 array of latent vectors is then flattened to a single vector of size 256 at the output to represent the graph embedding. 
This architecture is inspired by the transformer read-out layer proposed in ~\cite{jain2021representing} but modified to be similar to perciever IO's cross attention architecture \cite{jaegle2021perceiver} to gracefully scale computation and memory with number of nodes.
\footnote{The regular transformer read-out layer proposed in ~\cite{jain2021representing} has a memory and computation footprint of $O(N^2)$ (where $N$ is the number of nodes), but the cross attention layer proposed in ~\cite{jaegle2021perceiver} (and ours) scales linearly with $N$.}

\subsection{Evaluation Metric}
\label{sec:eval}
The prediction tasks presented in this paper are all regression tasks that require minimizing a mean square error loss of the predicted values w.r.t. the ground truth.
MSE however, is notoriously sensitive to outliers and sensitive to the range of the predicted output. Therefore, to get a better quantitative measure of performance, we propose measuring accuracy of correct predictions within a certain resolution.
For a given dataset $\{x_i, y_i\}_{i=1}^N$ let $\hat{y}_i$ be the predictions of the model. We define Acc@K as

\begin{equation}
    \small
    \text{Acc@K} = \frac{1}{N}\sum_{j=1}^N\mathbbm{1}(\vert y_j - \hat{y}_j \vert \leq \frac{\max_i y_i - \min_i y_i}{K}).
    \label{eq:acc}
\end{equation}

This metric, measures the ratio of predictions that are within a certain precision w.r.t to their ground truth value. For instance in the voltage prediction task (with a range of 1V), $K = 100$ sets the precision to $10 mV$, i.e. $1\%$ of the range.

\section{Experiments}




We perform an extensive empirical study to validate our hypothesis by focusing on the following questions:
(1) Can we successfully pretrain GNN architectures to regress the DC voltage of output nodes on a variety of circuits? 
(2) How does a pretrained network perform on predicting the output voltage nodes of an unseen circuit topology that is slightly different than what it has been pre-trained on?
(3) Can we fine-tune the pre-trained backbone to improve voltage prediction generalization on new unseen circuits?
(4) Can we re-use the learned representations after pre-training, to efficiently learn to predict new unseen circuit level tasks (e.g. a design metric of a circuit) via fine-tuning?
(5) Can we improve optimization sample efficiency of state of the art model-based circuit optimization tasks (e.g. ~\cite{hakhamaneshi2019bagnet})?


\subsection{Dataset and circuits under study}

We perform our experiments on two sets of circuits. First one is a family of resistor-based circuits that allow us more control over different generalization aspects of the problem. 
Second one is different variations of a real-world transistor-based two stage operational amplifier (OpAmp) circuit that illustrates the generalization of our method to more real-world examples where modeling design constraints is challenging.

\begin{figure}
    \centering
    \includegraphics[width=0.75\linewidth]{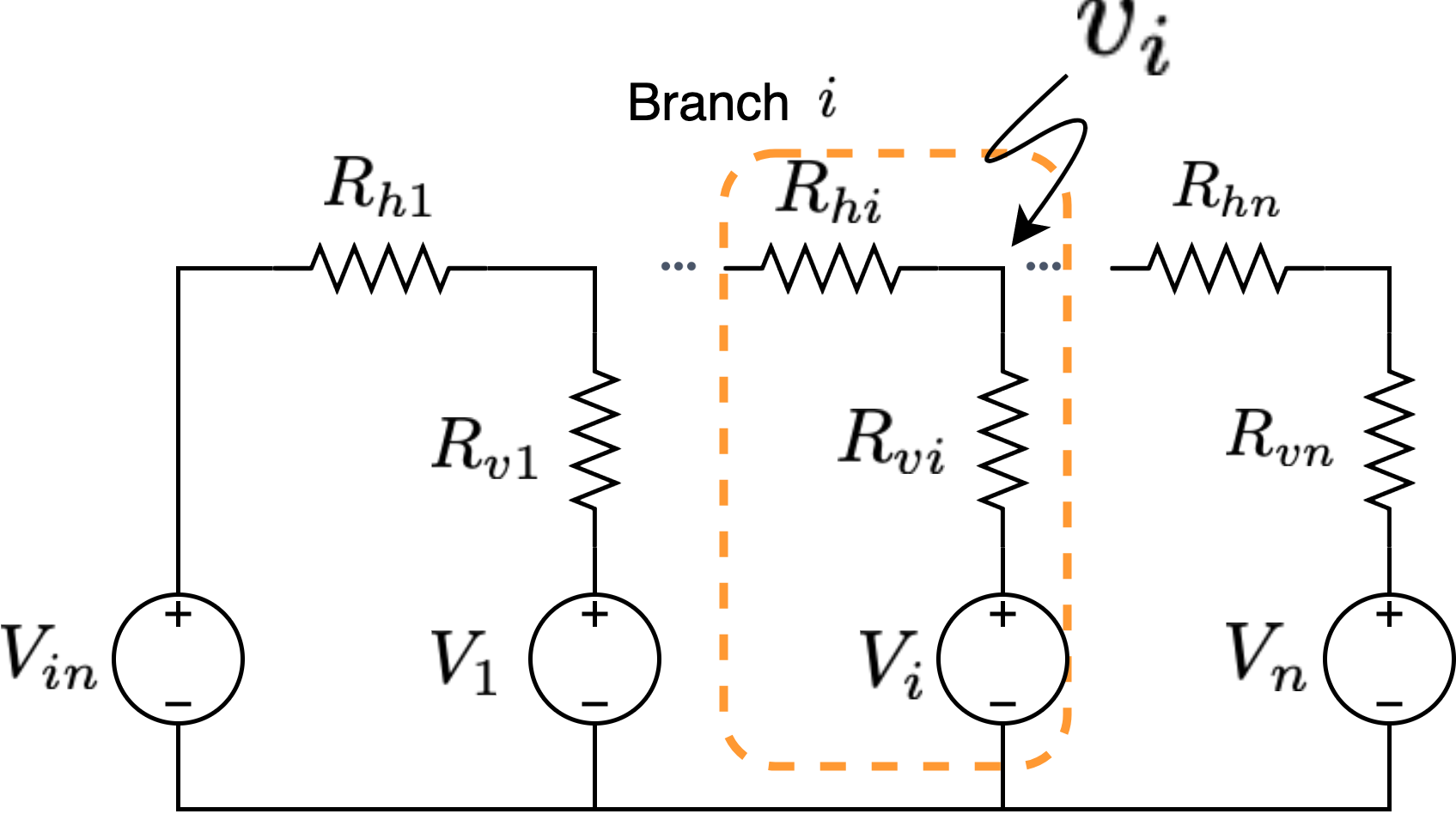}
    \small
    \caption{The circuit schematic of the resistor ladder. This circuit is comprised of $n$ branches. The pretraining task is to predict the output voltage of each branch $v_i$. A resistor ladder with $n$ branches has $2n$ resistors, $n+1$ voltage sources, and $n$ output nodes. }
    \label{fig:res-ladder}
\end{figure}

\textbf{Resistor-based circuits.} For pre-training we chose a particular topological structure made only of resistors and voltage sources known as resistor ladders (Figure \ref{fig:res-ladder}). 
For data for pre-training we can easily vary the number of ladder branches, values of resistors and voltage sources to cover enough support for resistors and voltage sources in various contextualized neighbourhoods in the graphs. 
During test we can see if the model has transferable domain knowledge relevant to resistor-based circuits by evaluating resistor ladders with more number of branches or new unseen resistor-voltage source topological structures. 
To this end, we generate 20k training instances of resistor ladders with 2 to 10 branches with equal distribution weight for each. For each node in each circuit we solve circuit equations to get the ground truth for voltage values.

\begin{figure}
    \centering
    \includegraphics[width=0.75\linewidth]{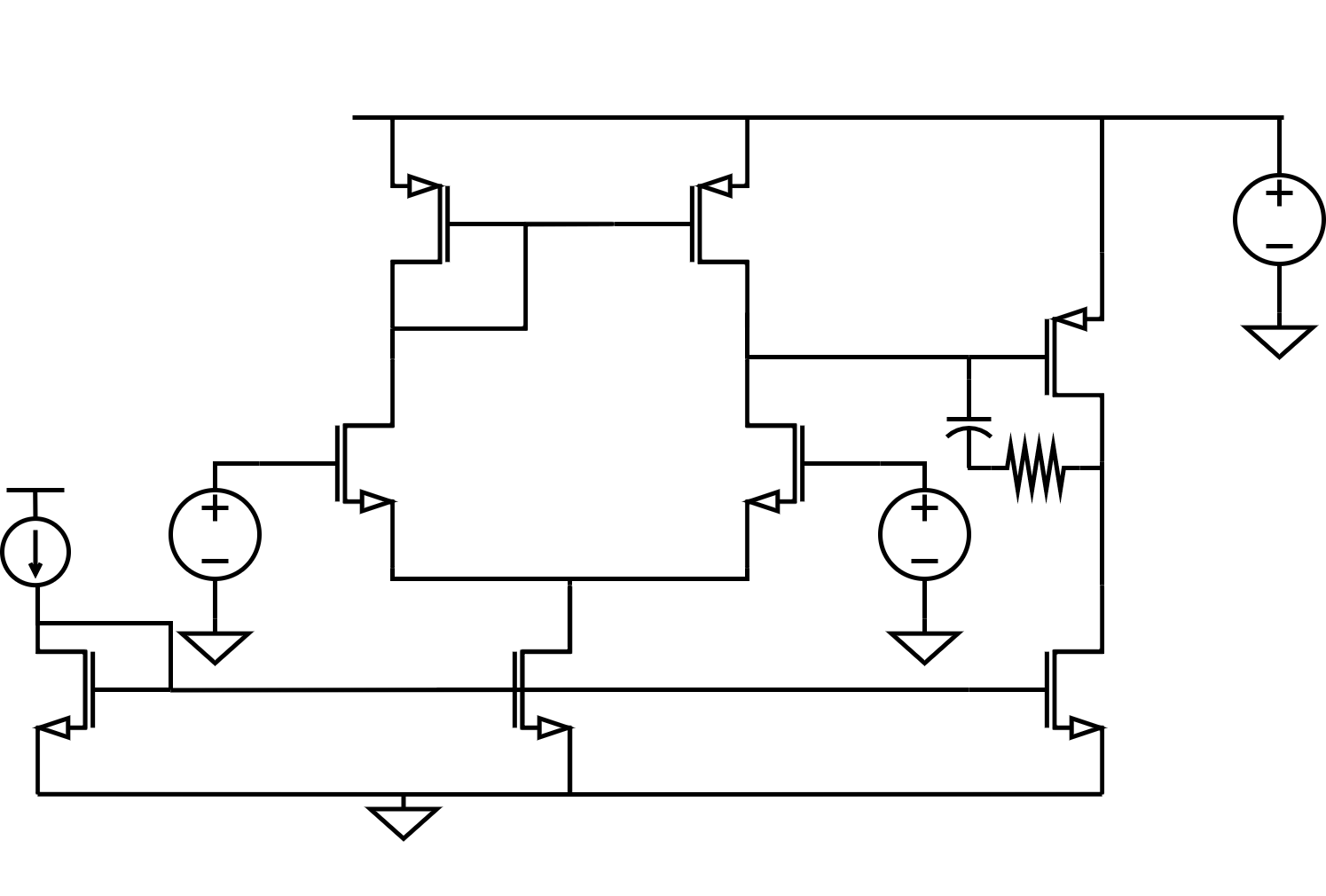}
    \small
    \caption{The circuit schematic of the pretraining dataset for the OpAmp experiments.}
    \label{fig:opamp}
\end{figure}

\textbf{Operational Amplifier (OpAmp).} This family of circuits has been studied in many prior work in circuit design optimization and modeling ~\cite{hakhamaneshi2021jumbo, settaluri2020autockt, hakhamaneshi2019bagnet, lyu2018batch}, due to its complexity in modeling real-world design constraints and existing trade-offs between different design choices. 
For pre-training we consider one variation of this circuit, shown in Figure \ref{fig:opamp}.
We generate 126 thousand design instances by varying device parameters, voltage and current source values, and the place that circuit input is applied (i.e. differential input, common mode input, supply input, .etc).
To get the ground truth we run Spice simulation on each instance and store all DC voltages along with the graph representation of the circuit. 
During test time, we evaluate our predictions on new topological modifications of this circuit where the current mirror self-bias connection is converted to a voltage-biased connection (Figure \ref{fig:opamp-biased-pmos}).
This minor topological modification induces enough behavioral change to the circuit that for proper generalization, the model has to have learned fine details on device-device interaction during pre-training.


\subsection{Neural network architecture design}
For all experiments we opted to use DeepGEN architecture proposed in ~\cite{li2020deepergcn} as the GNN backbone. 
This architecture has been proposed as a solution to train very deep Graph Convolution Networks.
It leverages generalized learnable aggregation layers as well as pre-activation skip connections to battle the vanishing gradient and over-smoothing problem often seen in making GNNs deep.
We found that this architecture generally outperforms other vanilla GNN architectures such as GCN \cite{kipf2016semi}, GAT \cite{velivckovic2017graph}.


\begin{table*}[tbh]
\centering
\caption{Test Acc@100 of prediction of voltages for each class of resistor ladders. We ran each experiment 3 times with random seeds for getting the standard errors.}
\label{table:main_results}
\begin{tabular}{@{}lllll@{}}
\toprule
           & R12                 & R16                 & R32                 & R128                \\ \midrule
MLP-5-fixed      & 0.3064 $\pm$ 0.0079 & 0.2466 $\pm$ 0.0056 & 0.1649 $\pm$ 0.0014 & 0.0905 $\pm$ 0.0009 \\ \midrule
DeepGEN-5-fixed  & 0.1024 $\pm$ 0.0073 & 0.1073 $\pm$ 0.0071 & 0.1114 $\pm$ 0.0084 & 0.1150 $\pm$ 0.0038 \\
DeepGEN-10-fixed & 0.6984 $\pm$ 0.0090 & 0.6585 $\pm$ 0.0110 & 0.6142 $\pm$ 0.0392 & 0.6167 $\pm$ 0.0027 \\
DeepGEN-15-fixed & 0.9124 $\pm$ 0.0112 & 0.8756 $\pm$ 0.0493 & 0.8756 $\pm$ 0.0166 & 0.8329 $\pm$ 0.0419 \\ \midrule
DeepGEN-5-(FPT)  & 0.1029 $\pm$ 0.0013 & 0.1038 $\pm$ 0.0026 & 0.1001 $\pm$ 0.0027 & 0.1009 $\pm$ 0.0027 \\
DeepGEN-10-(FPT) & 0.6847 $\pm$ 0.0069 & 0.6493 $\pm$ 0.0039 & 0.5979 $\pm$ 0.0064 & 0.5660 $\pm$ 0.0095 \\
DeepGEN-15-(FPT) & 0.8980 $\pm$ 0.0068 & 0.8711 $\pm$ 0.0059 & 0.8406 $\pm$ 0.0127 & 0.8186 $\pm$ 0.0093 \\ \bottomrule
\end{tabular}%
\end{table*}

\subsection{Do we learn to model DC voltages?}
\begin{figure}
    \centering
    \includegraphics[width=0.9\linewidth]{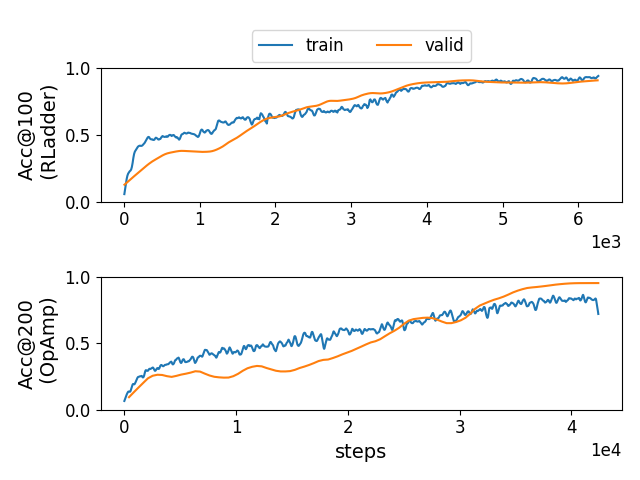}
    \small
    \caption{Train and Valid Accuracy of the model during pretraining on the resistor ladder circuit (top) and OpAmp circuit (bottom). Validation Acc@100 for resistor ladder reaches to 92\% and Acc@200 for OpAmp reaches to 91\% during pretraining.}
    \label{fig:train-curve}
\end{figure}

We pretrain the NN parameters with an Adam optimizer with a batch size of 256 until validation accuracy reaches its maximum. Figure \ref{fig:train-curve} shows the train and validation Acc@K for both resistor-ladder (top) and opamp (bottom) circuits. The validation set is simply a random split on the original dataset and it is clear that we can successfully reach to $> 90\%$ accuracy for both problems during pretraining. 

\subsection{Generalization to new unseen scenarios}

\subsubsection{Zero-shot generalization to new topologies}
\label{sec:zero-shot}

We first investigate whether the GNN backbone and the proceeding MLP (architecture of Figure \ref{fig:pretraining}) can generalize their prediction to new unseen topologies of resistor ladders with more number of branches and hence more complexity. To this end, we generate separate graph datasets for resistor ladders with 12, 16, 32, and 128 branches. A single graph example with 10 branches (the largest graph in the pretraining dataset) has 84 nodes, and 154 edges while with 128 branches it has 1032 nodes, and 2310 edges, an order of magnitude increase in graph complexity. For each circuit type we generate 1k circuit instances as test set and 20k other instances for training other baselines for comparison. We consider the following approaches as baselines: 

1. We take the pretrained model, freeze it and evaluate its prediction capability in a zero-shot manner with no finetuning. 
We refer to this approach as \textbf{FPT} (frozen pre-trained) model. In this modeling scheme, there is a topological mismatch between train and test sets.
The model only gets trained on graphs with 2 to 10 branches but is then evaluated on 12, 16, 32, and 128 branches which are more complex and require domain knowledge of how resistors and voltage sources behave in context of resistor ladders.

2. We use the 20k downstream training set to train a \textit{separate} predictive model, specialized for each branch size. In this modeling scheme, there is no mismatch between train and test sets. Note that the size of this dataset is similar to the size of pretraining dataset. For this strategy, since the input has a fixed static shape we can also represent it as a fixed size vector by concatenating the node features into a vector and use an MLP backbone as the node feature extractor. This is referred to as \textbf{MLP-5-fixed} since it has 5 layers. By comparing this model and GNNs we can see the impact of including the structural information in the model architecture. The specialized GNN models are referred to as \textbf{DeepGEN-x-fixed} in Table \ref{table:main_results} (where x is the number of layers). 


\textbf{Results.} Table \ref{table:main_results} compares the Acc@100 of the predictions of GNN backbones with different number of layers. All the models were trained with the same MSE loss, with batch size of 256 using Adam optimizer\footnote{Only for DeepGEN-15-fixed the batch size had to be reduced from default of 256 to 32 and 16 for R32 and R128, respectively just to prevent exceeding the GPU memory}. We first notice that the FPT model has an on par accuracy with the specialized models that were separately trained on comparable dataset size from a single topology. This result is despite the fact that FPT models were never trained on the test topologies. 
Moreover, we also note that deeper models have better generalization across the board. This could be due to lack of large enough receptive field over the entire graph for capturing long range dependencies in shallower models.

Also, as evident by the poor performance of the results of MLP-5-fixed in Table \ref{table:main_results}, we conclude that 
the structural information built into the model architecture of GNNs can significantly improve the generalization of the model. This is an important evidence for choosing GNNs over MLPs even in a fixed topology setup since they have a better inductive bias for modeling circuits. 


\begin{figure*}
    \centering
    \includegraphics[width=0.85\textwidth]{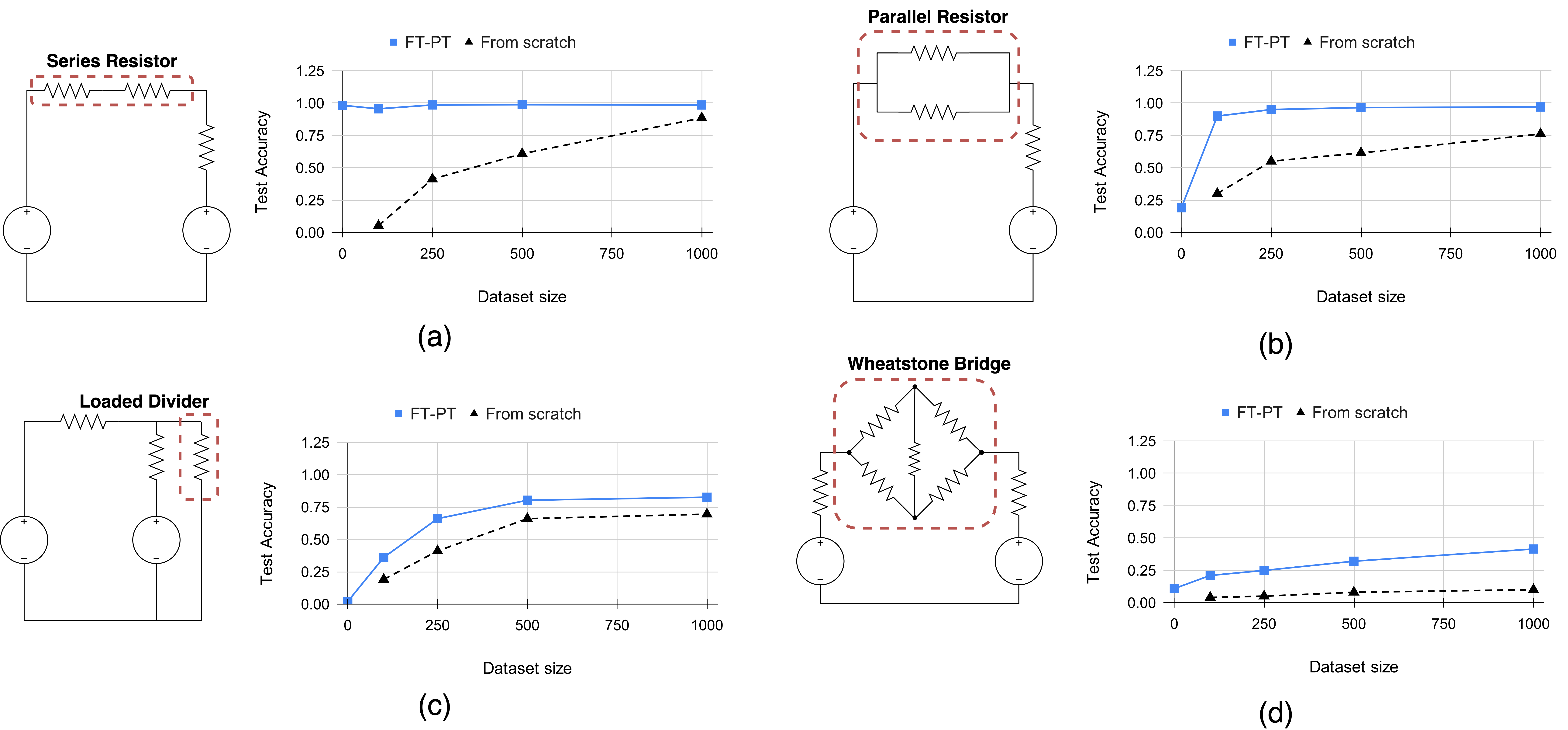}
    \small
    \caption{Finetuning capabilities of the pretrained model to new unseen topologies comprised of resistors and voltages sources. Each figure shows models that are trained and evaluated on the illustrated resistor networks where new unseen topological configurations are highlighted in red dashed boxes.}
    \label{fig:fine-tuning results}
\end{figure*}

\begin{figure*}[t]
    \centering
    \begin{subfigure}{0.32\textwidth}
        \includegraphics[width=\textwidth]{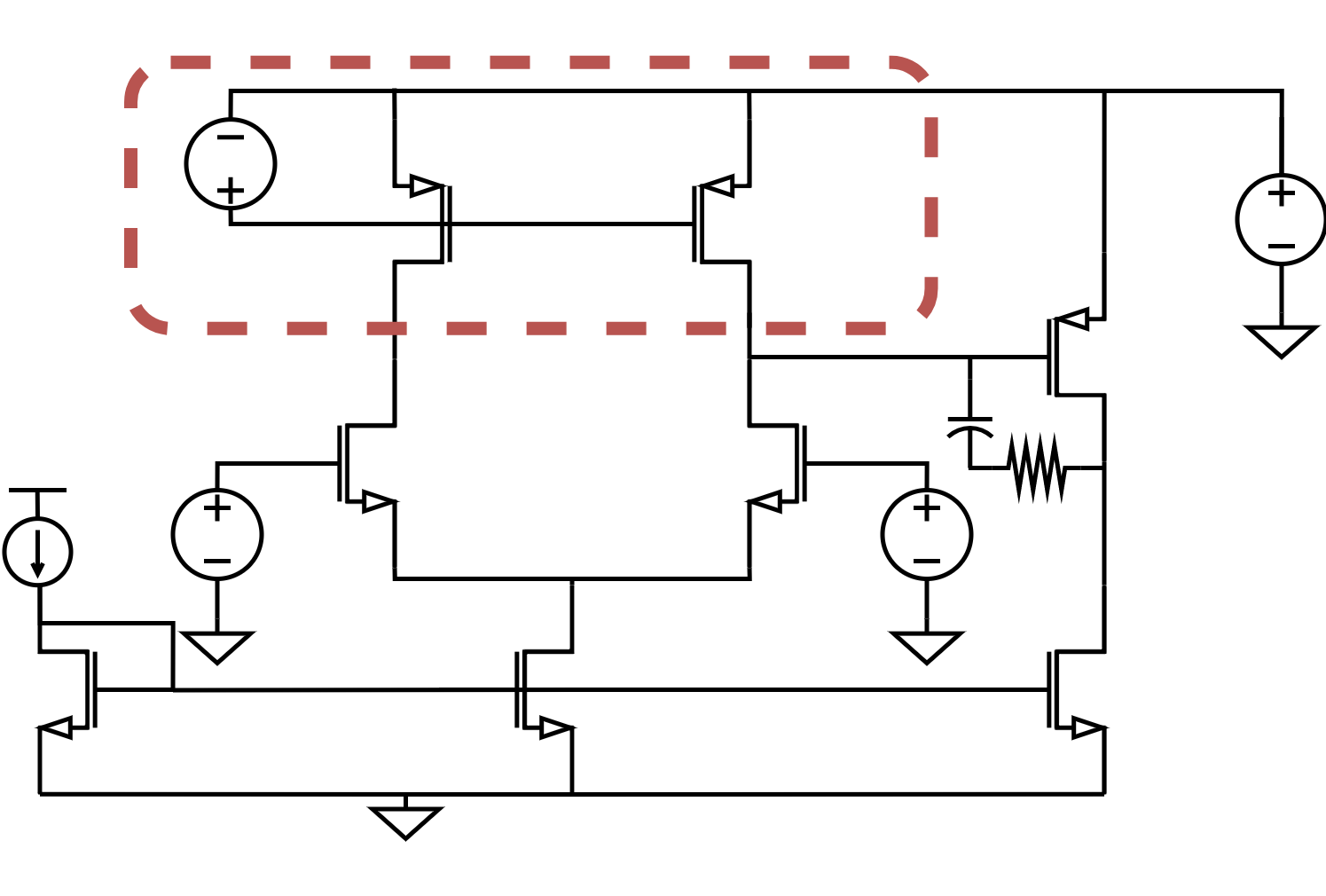}\par
        \caption{}\label{fig:opamp-biased-pmos}
    \end{subfigure}
    \hfill
    \begin{subfigure}{0.32\textwidth}
            \includegraphics[width=\textwidth]{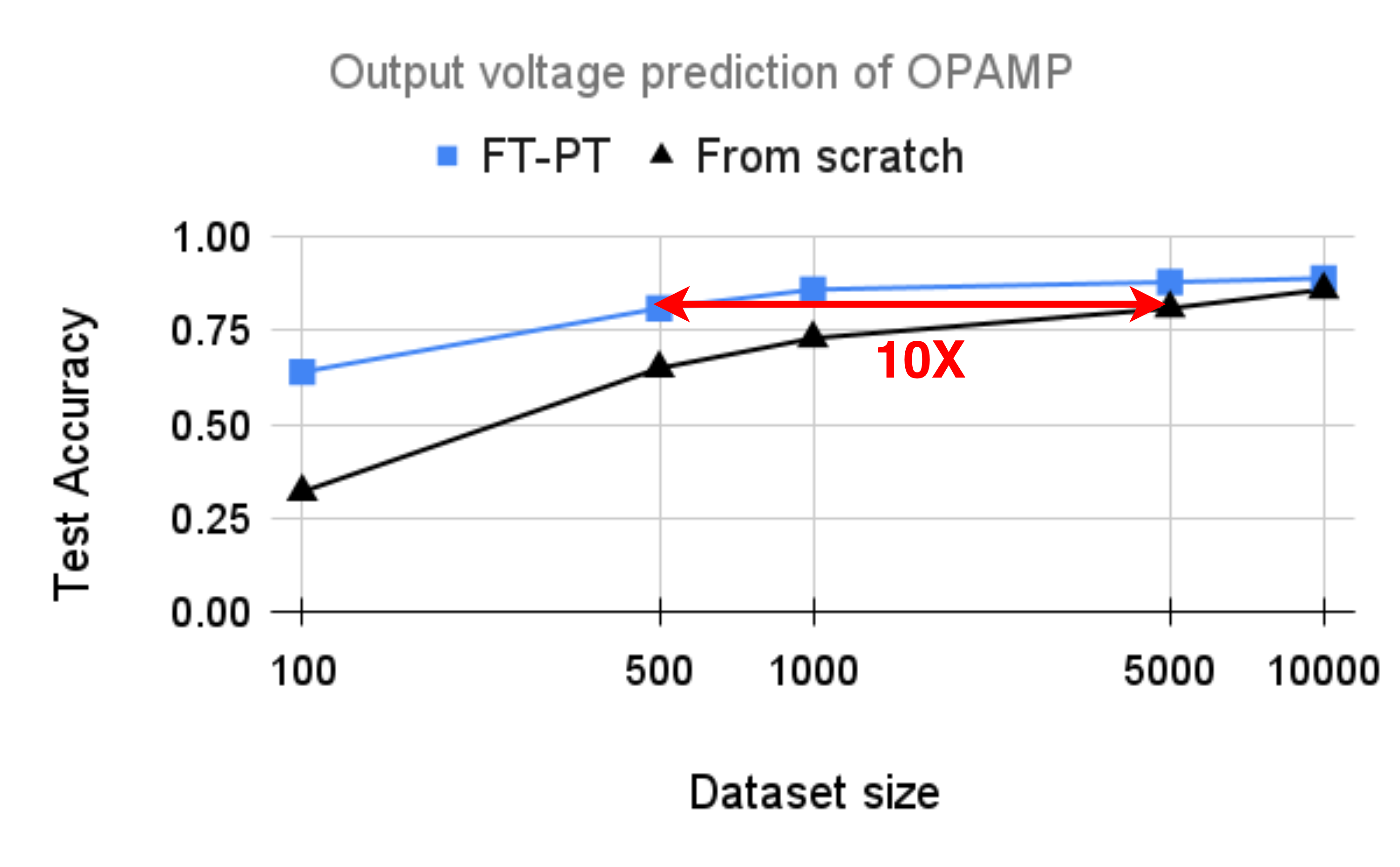}\par
            \caption{}\label{fig:opamp-dc}
    \end{subfigure} 
        \hfill
    \begin{subfigure}{0.32\textwidth}
        \includegraphics[width=\textwidth]{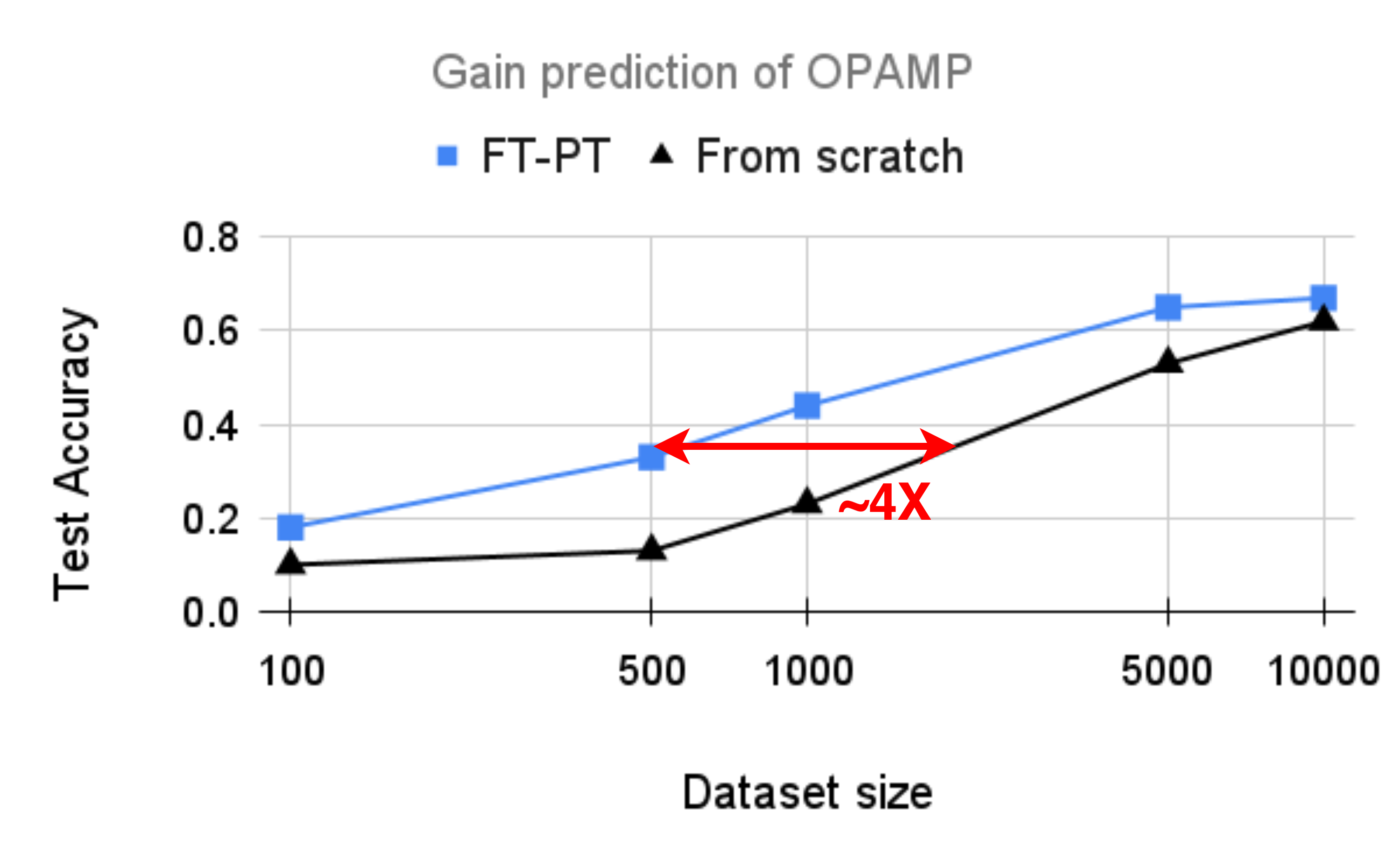}\par
        \caption{}\label{fig:opamp-gain}
    \end{subfigure}
    \caption{
    (a) The unseen topology of the two-stage opamp that the GNN gets fine-tuned on. We collect a small dataset of 10 thousand designs from this topology along with their ground truth output values. We then use fractions of this training dataset to show how much improvements we get from more data.
    (b) shows the test Acc@200 of predicting the output voltage for our fine-tuned method (FT-PT) vs. training from scratch with no knowledge transfer. (c) shows the test Acc@50 comparison for the gain prediction task (i.e. also a new graph property prediction task).
    }
\end{figure*}

\subsubsection{Few-shot generalization to other unseen topologies}
\label{sec:few-shot}



In this experiment, we evaluate the FPT models on circuits that are slightly more different than the circuits that they have been pretrained on (e.g. resistor divider with parallel or series resistor combinations in case of a the resistor ladder pretraining dataset).
These small modifications, which are highlighted with dashed red lines in Figure \ref{fig:fine-tuning results} and \ref{fig:opamp-biased-pmos}, will result in poor zero-shot performance, but we show that we can significantly improve the prediction performance via finetuning on small datasets from these topologies. 
For resistor-based circuits we generate 1k graphs per each topology for training and then a separate 1k instance held-out dataset for evaluation. For the OpAmp we generate 10k samples for training and 1k examples for evaluation from the topology shown in Figure \ref{fig:opamp-biased-pmos} in which the self-biased current mirror connection is removed and the gate connections are replaced by a voltage source. For baselines, we compare the finetuned models (FT-PT) against training the same architecture from scratch. For the remainder of experiments we use DeepGEN-15 as the GNN backbone unless otherwise mentioned.


\textbf{Results.} Figure \ref{fig:fine-tuning results} and \ref{fig:opamp-dc} show this comparison for each circuit. For resistive circuits we finetune the backbone on 10, 25, 50, and 100\% of the training dataset and for the OpAmp we finetune with 1, 5, 10, 50, and 100 \% of the dataset. We can see that even with a severe distribution shift from pretraining and test domains, finetuning the learned representation can be more efficient than learning from scratch. 

\subsection{Domain knowledge transfer by re-using the learned GNN features for predicting graph level tasks}

In this section we use the architecture outlined in Figure \ref{sec:node-to-graph} to re-use the learned node features of the pretrained GNN by using a cross attention pooling layer to construct a graph feature from aggregating node features and use that for prediction of graph-level metrics. To learn the model, we update all parameters (including the GNN backbone parameters) using Adam optimizer on the downstream training dataset. 

For this experiment, we have generated 10k instances of circuits from Figure \ref{fig:opamp-biased-pmos} that have ground truth values of the gain of the amplifiers which are collected by running simulations. We then evaluate the generalization of our learned models (trained by fine-tuning or from scratch) on 1k separate held-out dataset collected in the same procedure.


\textbf{Results.} 
Figure \ref{fig:opamp-gain} shows the accuracy of gain prediction after fine-tuning on various dataset sizes. As illustrated in this figure we can see that the pretrained features can significantly improve generalization capabilities of the model even with small fractions of the training data.

\begin{figure}
    \centering
    \includegraphics[width=\linewidth]{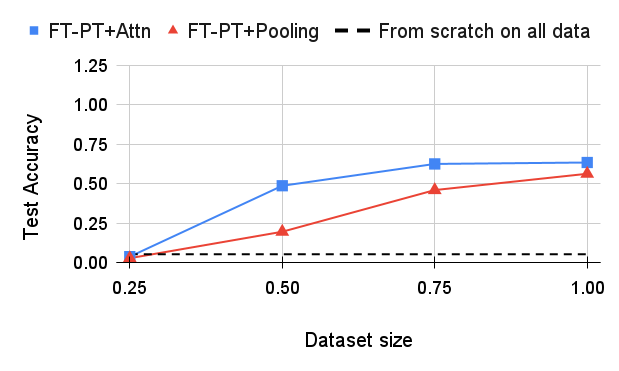}
    \small
    \caption{Acc@100 of the output resistance prediction task on resistor ladder. 
    Training from scratch will fail to generalize to the test set even if all 1k data points are used due to significant over-fitting. Moreover the attention-based aggregation layer outperforms the naive average pooling aggregation.}
    \label{fig:rout-pred}
\end{figure}

To study how important the choice of using attention-based pooling layer is over simpler alternative pooling methods, we have also conducted an ablation experiment where we compare our method against a simple average pooling of all node features to construct the graph feature. We perform this ablation, on the task of predicting the output resistance of the resistor ladders with 10 branches. Figure \ref{fig:rout-pred} compares these baselines. As seen in the figure, our method outperforms the simple average pooling method usually used in literature \cite{hu2020open}. We can also see that pre-training the GNNs prevent the model from overfitting to the small available downstream dataset as it does if trained from scratch.


\subsection{Improving sample efficiency of model-based optimization by using pre-trained models}

In this section, we show the benefits of pre-trained circuit models in improving sample efficiency of model-based circuit optimization algorithms. 
In these methods, part of the objective is to learn a good model of the design objectives so that we can compare different design choices and focus exploration on more promising regions.
To this end, we consider a simple and yet powerful instantiation of this method introduced in BagNet \cite{hakhamaneshi2019bagnet} that uses a learned discriminator neural network to compare designs during the evolutionary optimization procedure to avoid running unnecessary simulations to save optimization time.

\begin{figure}
    \centering
    \includegraphics[width=\linewidth]{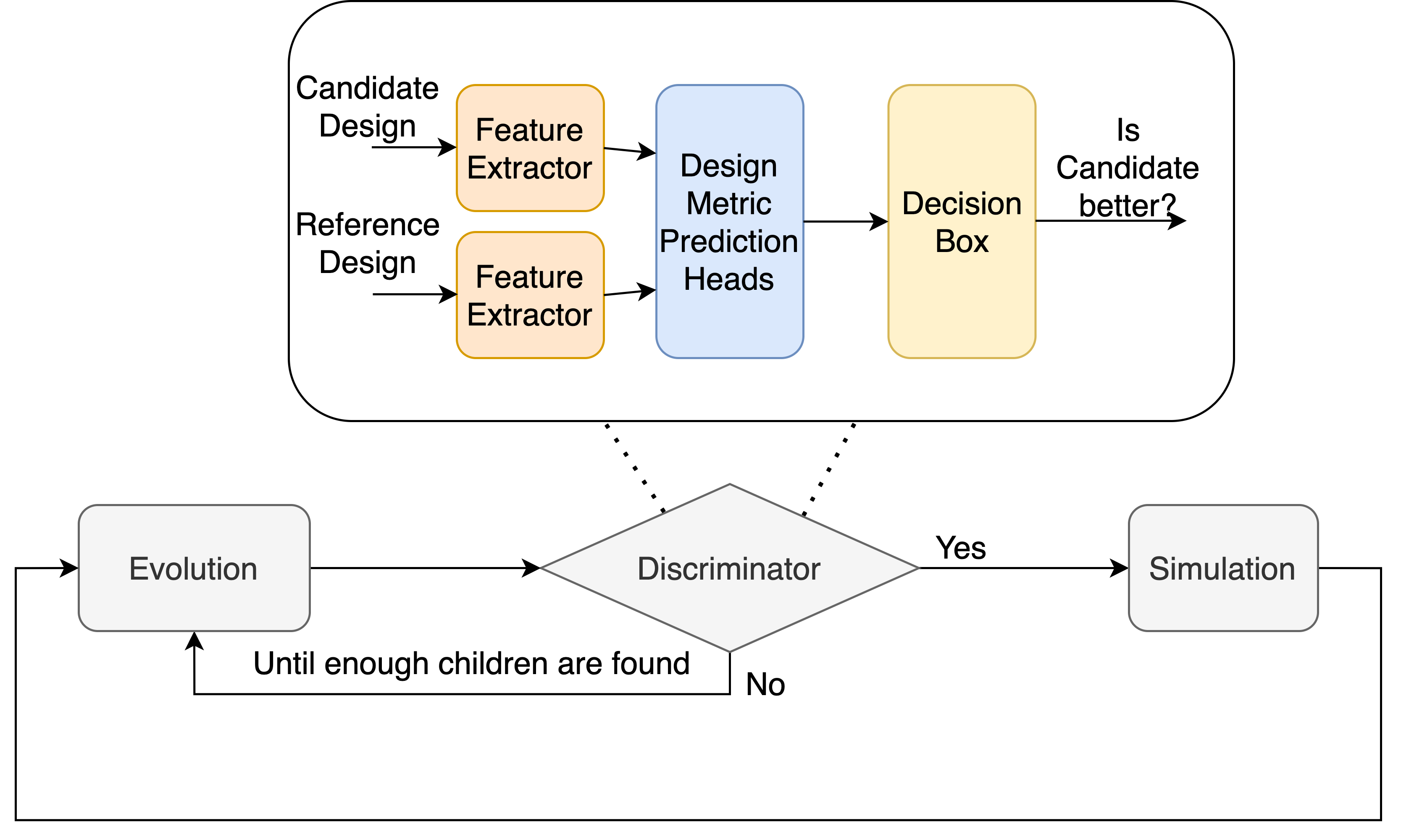}
    \small
    \caption{An overview of the BagNet algorithm. It uses Neural network discriminator during the optimization to pre-select promising designs before performing simulation. This way simulation is only done on designs that will most likely result in improvement of the objective.}
    \label{fig:bagnet}
\end{figure}

Figure \ref{fig:bagnet} shows the high-level overview of the method. In each iteration the new generated population is evaluated via the learned NN discriminator, comparing each individual new candidate against other designs that have already been simulated, with known performance.
In the original work, a simple multi-layer perceptron was used as the backbone feature extractor which was shown to out-perform the base evolutionary algorithm with no such discriminator.
\footnote{For more information on the method we refer the reader to the original paper.}

In this section we reproduce those results and show that if we use the pre-trained GNN backbone for extracting circuit features we can significantly boost the performance to almost as good as an oracle discriminator which has access to the ground truth design metrics and their comparisons. 
To validate our hypothesis, we use the Operational Amplifier introduced in Figure \ref{fig:opamp} as our pretraining topology. 
In this circuit the compensation method is comprised of  a series capacitor and resistor ($C_c-R_z$). However, for optimization we consider two target topologies: one is same as the pre-training topology and another one with a different topology where only a capacitor is used for compensation ($C_c$). 
It is known that adding a resistor can make the compensation easier to expand the bandwidth of the amplifier. We define the same design constraints as in the original BagNet paper for both circuits. 

We consider the following baselines: \textbf{Evolutionary (Evo)} just runs the base evolutionary algorithm and does not use any discriminator during the optimization, \textbf{Oracle} uses ground truth comparison labels for the discriminator, \textbf{BagNet + FC} uses fully-connected layers as feature extractor (the original BagNet work), \textbf{BagNet + Randinit GNN}, uses a randomly initialized GNN as feature extractor that is jointly trained with the the rest of the model during optimization, \textbf{BagNet + FPT GNN} uses a frozen but pre-trained GNN for feature extractor, and \textbf{BagNet + FT-PT GNN} fine-tunes the pretrained GNN.

\begin{figure}
    \centering
    \begin{subfigure}{\linewidth}
        \includegraphics[width=\linewidth]{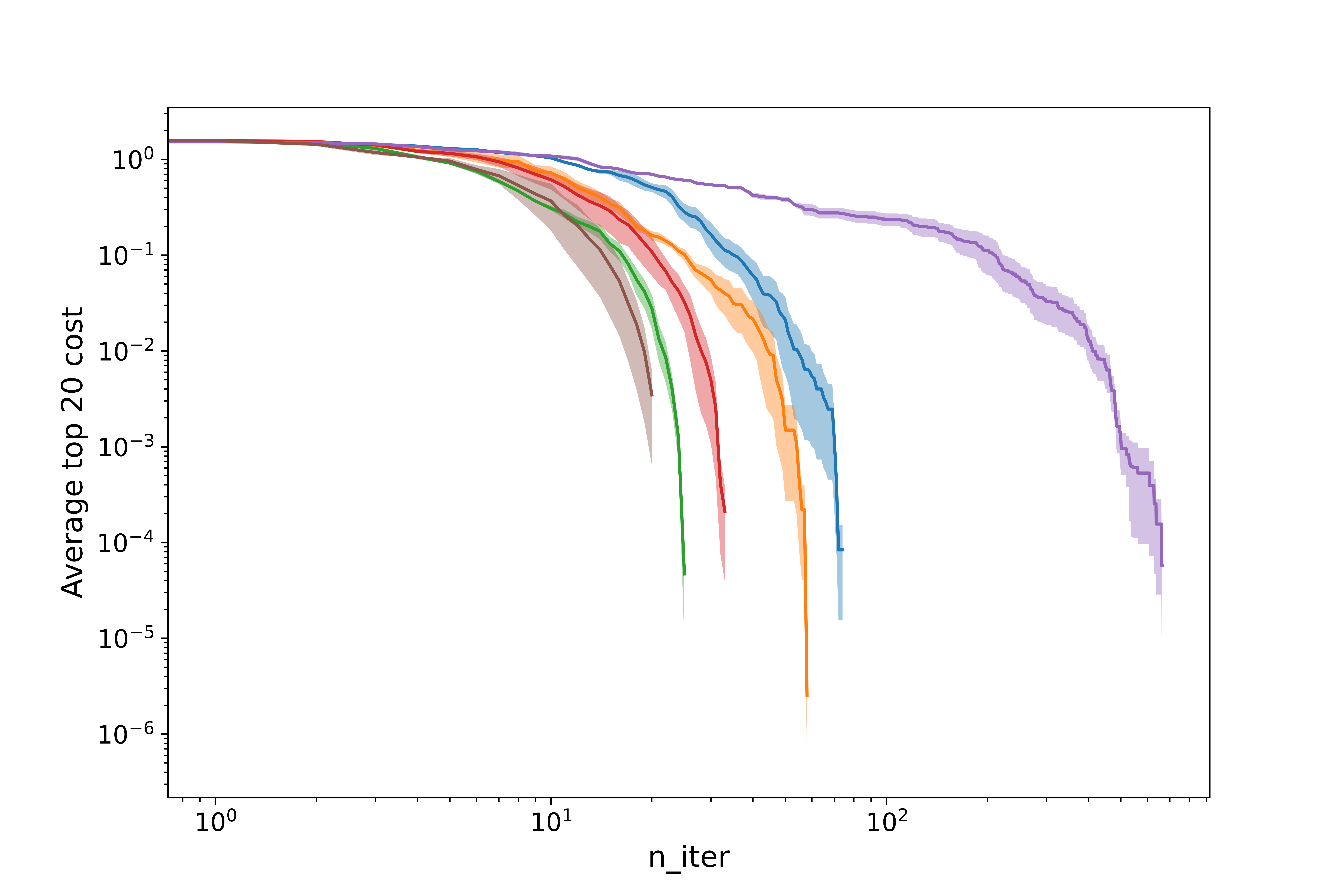}\par
    \end{subfigure}
    \par
    \begin{subfigure}{\linewidth}
        \includegraphics[width=\linewidth]{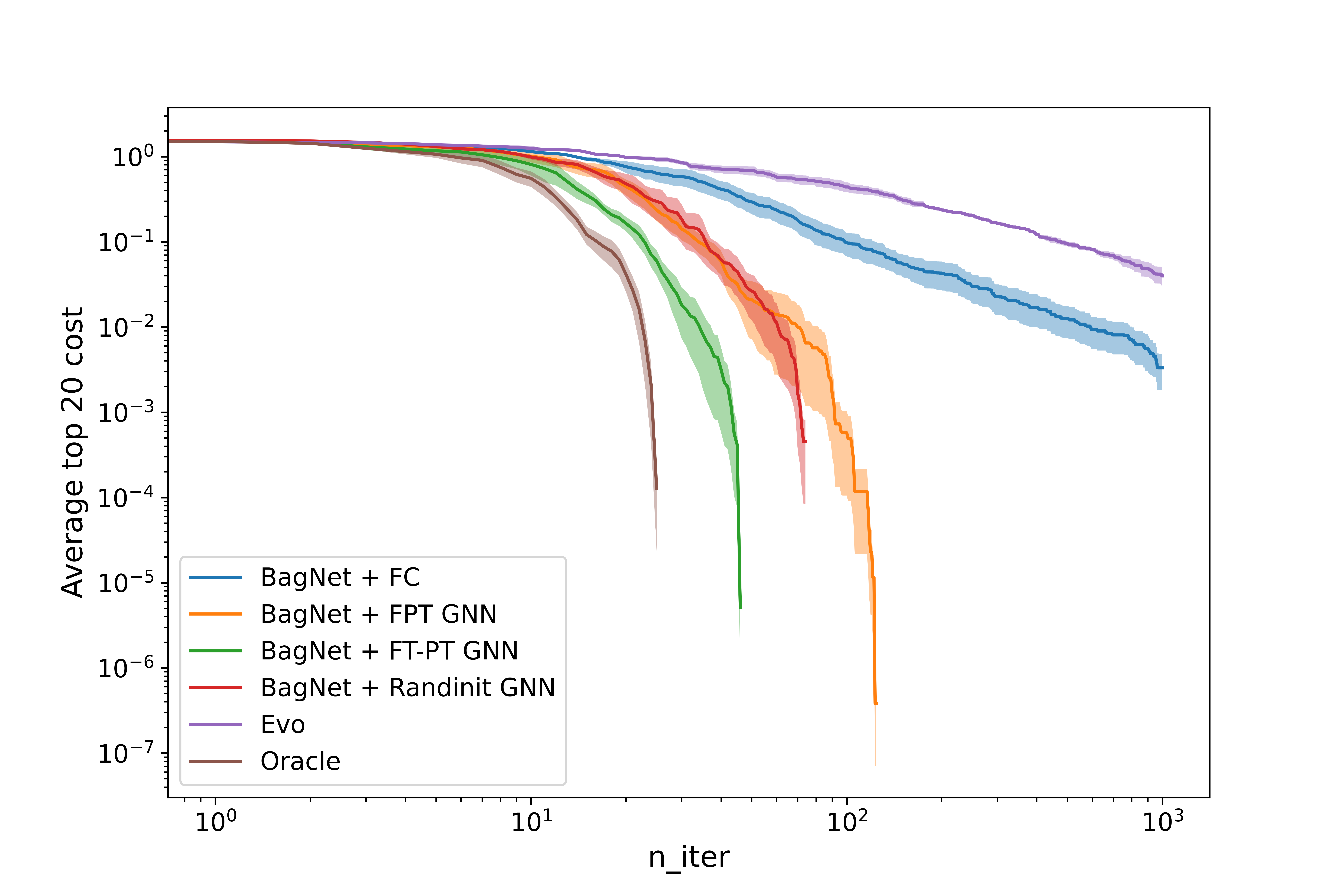}\par
    \end{subfigure}
    \small
    \caption{The average cost of top 20 designs over the course of optimization using different evolutionary-based approaches. (Top) The optimization is done on the OpAmp with $C_c-R_z$ compensation (Fig. \ref{fig:opamp}, (Bottom) The target topology is the OpAmp with only capacitor compensation ($C_c$)} 
    \label{fig:bagnet-res}
\end{figure}

\textbf{Results.} Figure \ref{fig:bagnet-res} qualitatively shows the quality of the designs found during optimization as the function of number of iterations. 
We run each algorithm with three initial seeds to show the standard error on performance. 
The y-axis measures the objective value of the top 20 designs found until each iteration and reaching a value of zero implies having found at least 20 designs that satisfy all design objectives. 
We also provide the average number of simulations and average number of discriminator queries for all of these baselines in Table \ref{tab:bagnet-res} for quantitative comparisons. 

All BagNet algorithms that use the discriminator outperform the base evolutionary algorithm. 
We note that the performance boost obtained by the pre-trained and fine-tuned GNN (FT-PT) is very close to the oracle baseline, while using the frozen pre-trained features or randomly initialized GNN is not as good. This gap is larger when tried on the harder optimization problem ($C_c$-only), despite pretraining FT-PT's initial weights on a different topology than the target optimization topology.
This picture clearly shows that the performance boost in sample efficiency is partly due to the utilization of GNNs over MLP architectures as feature extractors and partly due to being able to transfer the pretrained representations to unseen topologies. 
We also note that the number of discriminator queries are typically lower when the model is better. 
This implies that the more accurate the model gets the number of new candidate designs that get rejected by mistake become lower.

\begin{table}[]
\centering
\small
\caption{The comparison of sample complexity of the optimization algorithms.}
\label{tab:bagnet-res}
\resizebox{\linewidth}{!}{%
\begin{tabular}{lllll}
                                   & \multicolumn{2}{c}{$C_c-R_z$}    & \multicolumn{2}{c}{$C_c-only$}    \\ \hline
                                   & Queries         & Sims           & Queries          & Sims           \\ \hline
Oracle                             & 3770.7          & 193.0          & 18821.0          & 224.3          \\
Evo                                & -               & 2757.3         & -                & 5422.7         \\
BagNet + FC                        & 14720.0         & 397.7          & 176568.3         & 4024.0         \\
BagNet + Randinit GNN              & 8892.0          & 254.7          & 47961.7          & 426.5          \\
BagNet + FPT GNN                   & 8000.3          & 347.3          & 27309.0          & 500.3          \\
\textbf{BagNet + FT-PT GNN (Ours)} & \textbf{4890.7} & \textbf{226.0} & \textbf{14408.7} & \textbf{293.0} \\ \hline
\end{tabular}%
}
\end{table}

\section{Conclusion}


This work illustrates how we can use large in-expensive datasets collected from running DC simulations at scale on analog circuits to learn transferable circuit representations that could enable more sample efficient modeling and optimization of new expensive-to-collect metrics. 

We presented a supervised-learning paradigm for learning representations that are transferable to new circuit topologies that can be fine-tuned for new graph level prediction tasks.
We introduced a novel graph representation of the circuit and used the prediction of the output node voltages as the pretraining objective.
We showed that such a pretraining objective can induce learning generalizable domain specific features that can be easily fine-tuned for similar predictions on new topologies or unseen new graph property prediction tasks such as predicting the output resistance of a circuit or gain of an amplifier. 
We have also showed that we can leverage the pre-trained features to boost the performance of model-based optimization methods used for analog circuit design automation.

\section{Acknowledgement}

This work is supported in part by NSF AI Institute for Advances in Optimization (AI4OPT), Berkeley Wireless Research Center (BWRC) and Berkeley AI Research (BAIR) member companies.

\appendices


\section{Hyper-parameter and compute details}

All the models are trained by minimizing the mean square error loss on the predicted outputs using Adam optimizer and learning rate with linear decay from $1e-3$ to $5e-5$. The other common hyper-paramters are batch size of 256, activation of Relu, and hidden channel size of 128 for GNNs and 512 for MLPs. The longest training run-time belongs to running DeepGEN-15 which approximately takes $\sim5$ hrs on a single NVIDIA GeForce GTX TITAN X.

\section{Access to code and environment}
The code for datasets, data loaders, and evaluators are released under BSD 3-Clause license at \href{https://github.com/kouroshHakha/circuit-fewshot-code}{https://github.com/kouroshHakha/circuit-fewshot-code}. The data loaders are compatible with Pytorch Geometric \cite{fey2019fast} interface. We provide automatic dataset downloading, processing, and evaluation, as well as the implementation of example baselines discussed in the paper. We have a separate repository for data generation through circuit simulators that can be found at \href{https://github.com/kouroshHakha/circuit-fewshot-data}{https://github.com/kouroshHakha/circuit-fewshot-data}. The raw datasets generated from this code-base are automatically downloaded when the previous code-base is utilized. 

The code for optimization experiments is found under \texttt{bagnet\_gnn} branch of the existing repository of BagNet at \href{https://github.com/kouroshHakha/bagnet_ngspice/tree/bagnet_gnn}{https://github.com/kouroshHakha/bagnet\_ngspice/tree/bagnet\_gnn}.

\newpage
\bibliography{IEEEabrv, main}

\begin{thebibliography}{10}
\providecommand{\url}[1]{#1}
\csname url@samestyle\endcsname
\providecommand{\newblock}{\relax}
\providecommand{\bibinfo}[2]{#2}
\providecommand{\BIBentrySTDinterwordspacing}{\spaceskip=0pt\relax}
\providecommand{\BIBentryALTinterwordstretchfactor}{4}
\providecommand{\BIBentryALTinterwordspacing}{\spaceskip=\fontdimen2\font plus
\BIBentryALTinterwordstretchfactor\fontdimen3\font minus
  \fontdimen4\font\relax}
\providecommand{\BIBforeignlanguage}[2]{{%
\expandafter\ifx\csname l@#1\endcsname\relax
\typeout{** WARNING: IEEEtran.bst: No hyphenation pattern has been}%
\typeout{** loaded for the language `#1'. Using the pattern for}%
\typeout{** the default language instead.}%
\else
\language=\csname l@#1\endcsname
\fi
#2}}
\providecommand{\BIBdecl}{\relax}
\BIBdecl

\bibitem{makrani2019xppe}
H.~M. Makrani, H.~Sayadi, T.~Mohsenin, S.~Rafatirad, A.~Sasan, and H.~Homayoun,
  ``Xppe: cross-platform performance estimation of hardware accelerators using
  machine learning,'' in \emph{Proceedings of the 24th Asia and South Pacific
  Design Automation Conference}, 2019, pp. 727--732.

\bibitem{haaswijk2018deep}
W.~Haaswijk, E.~Collins, B.~Seguin, M.~Soeken, F.~Kaplan, S.~S{\"u}sstrunk, and
  G.~De~Micheli, ``Deep learning for logic optimization algorithms,'' in
  \emph{2018 IEEE International Symposium on Circuits and Systems
  (ISCAS)}.\hskip 1em plus 0.5em minus 0.4em\relax IEEE, 2018, pp. 1--4.

\bibitem{hosny2020drills}
A.~Hosny, S.~Hashemi, M.~Shalan, and S.~Reda, ``Drills: Deep reinforcement
  learning for logic synthesis,'' in \emph{2020 25th Asia and South Pacific
  Design Automation Conference (ASP-DAC)}.\hskip 1em plus 0.5em minus
  0.4em\relax IEEE, 2020, pp. 581--586.

\bibitem{mirhoseini2021graph}
A.~Mirhoseini, A.~Goldie, M.~Yazgan, J.~W. Jiang, E.~Songhori, S.~Wang, Y.-J.
  Lee, E.~Johnson, O.~Pathak, A.~Nazi \emph{et~al.}, ``A graph placement
  methodology for fast chip design,'' \emph{Nature}, vol. 594, no. 7862, pp.
  207--212, 2021.

\bibitem{lu2019gan}
Y.-C. Lu, J.~Lee, A.~Agnesina, K.~Samadi, and S.~K. Lim, ``Gan-cts: A
  generative adversarial framework for clock tree prediction and
  optimization,'' in \emph{2019 IEEE/ACM International Conference on
  Computer-Aided Design (ICCAD)}.\hskip 1em plus 0.5em minus 0.4em\relax IEEE,
  2019, pp. 1--8.

\bibitem{xie2018routenet}
Z.~Xie, Y.-H. Huang, G.-Q. Fang, H.~Ren, S.-Y. Fang, Y.~Chen, and J.~Hu,
  ``Routenet: Routability prediction for mixed-size designs using convolutional
  neural network,'' in \emph{2018 IEEE/ACM International Conference on
  Computer-Aided Design (ICCAD)}.\hskip 1em plus 0.5em minus 0.4em\relax IEEE,
  2018, pp. 1--8.

\bibitem{wang2020gcn}
H.~Wang, K.~Wang, J.~Yang, L.~Shen, N.~Sun, H.-S. Lee, and S.~Han, ``Gcn-rl
  circuit designer: Transferable transistor sizing with graph neural networks
  and reinforcement learning,'' in \emph{2020 57th ACM/IEEE Design Automation
  Conference (DAC)}.\hskip 1em plus 0.5em minus 0.4em\relax IEEE, 2020, pp.
  1--6.

\bibitem{settaluri2020autockt}
K.~Settaluri, A.~Haj-Ali, Q.~Huang, K.~Hakhamaneshi, and B.~Nikolic, ``Autockt:
  deep reinforcement learning of analog circuit designs,'' in \emph{2020
  Design, Automation \& Test in Europe Conference \& Exhibition (DATE)}.\hskip
  1em plus 0.5em minus 0.4em\relax IEEE, 2020, pp. 490--495.

\bibitem{hakhamaneshi2019bagnet}
K.~Hakhamaneshi, N.~Werblun, P.~Abbeel, and V.~Stojanovi{\'c}, ``Bagnet:
  Berkeley analog generator with layout optimizer boosted with deep neural
  networks,'' in \emph{2019 IEEE/ACM International Conference on Computer-Aided
  Design (ICCAD)}.\hskip 1em plus 0.5em minus 0.4em\relax IEEE, 2019, pp. 1--8.

\bibitem{li2020customized}
Y.~Li, Y.~Lin, M.~Madhusudan, A.~Sharma, W.~Xu, S.~S. Sapatnekar, R.~Harjani,
  and J.~Hu, ``A customized graph neural network model for guiding analog ic
  placement,'' in \emph{2020 IEEE/ACM International Conference On Computer
  Aided Design (ICCAD)}.\hskip 1em plus 0.5em minus 0.4em\relax IEEE, 2020, pp.
  1--9.

\bibitem{jiang2020efficient}
Y.~Jiang, F.~Yang, B.~Yu, D.~Zhou, and X.~Zeng, ``Efficient layout hotspot
  detection via binarized residual neural network ensemble,'' \emph{IEEE
  Transactions on Computer-Aided Design of Integrated Circuits and Systems},
  vol.~40, no.~7, pp. 1476--1488, 2020.

\bibitem{ye2019lithogan}
W.~Ye, M.~B. Alawieh, Y.~Lin, and D.~Z. Pan, ``Lithogan: End-to-end lithography
  modeling with generative adversarial networks,'' in \emph{2019 56th ACM/IEEE
  Design Automation Conference (DAC)}.\hskip 1em plus 0.5em minus 0.4em\relax
  IEEE, 2019, pp. 1--6.

\bibitem{huang2021machine}
G.~Huang, J.~Hu, Y.~He, J.~Liu, M.~Ma, Z.~Shen, J.~Wu, Y.~Xu, H.~Zhang,
  K.~Zhong \emph{et~al.}, ``Machine learning for electronic design automation:
  A survey,'' \emph{ACM Transactions on Design Automation of Electronic Systems
  (TODAES)}, vol.~26, no.~5, pp. 1--46, 2021.

\bibitem{hassanpourghadi2021circuit}
M.~Hassanpourghadi, S.~Su, R.~A. Rasul, J.~Liu, Q.~Zhang, and M.~S.-W. Chen,
  ``Circuit connectivity inspired neural network for analog mixed-signal
  functional modeling,'' in \emph{2021 58th ACM/IEEE Design Automation
  Conference (DAC)}.\hskip 1em plus 0.5em minus 0.4em\relax IEEE, 2021, pp.
  505--510.

\bibitem{hakhamaneshi2021jumbo}
K.~Hakhamaneshi, P.~Abbeel, V.~Stojanovic, and A.~Grover, ``Jumbo: Scalable
  multi-task bayesian optimization using offline data,'' \emph{arXiv preprint
  arXiv:2106.00942}, 2021.

\bibitem{lyu2018batch}
W.~Lyu, F.~Yang, C.~Yan, D.~Zhou, and X.~Zeng, ``Batch bayesian optimization
  via multi-objective acquisition ensemble for automated analog circuit
  design,'' in \emph{International conference on machine learning}.\hskip 1em
  plus 0.5em minus 0.4em\relax PMLR, 2018, pp. 3306--3314.

\bibitem{ren2020paragraph}
H.~Ren, G.~F. Kokai, W.~J. Turner, and T.-S. Ku, ``Paragraph: Layout parasitics
  and device parameter prediction using graph neural networks,'' in \emph{2020
  57th ACM/IEEE Design Automation Conference (DAC)}.\hskip 1em plus 0.5em minus
  0.4em\relax IEEE, 2020, pp. 1--6.

\bibitem{ma2019high}
Y.~Ma, H.~Ren, B.~Khailany, H.~Sikka, L.~Luo, K.~Natarajan, and B.~Yu, ``High
  performance graph convolutional networks with applications in testability
  analysis,'' in \emph{Proceedings of the 56th Annual Design Automation
  Conference 2019}, 2019, pp. 1--6.

\bibitem{wolfe2003extraction}
G.~Wolfe and R.~Vemuri, ``Extraction and use of neural network models in
  automated synthesis of operational amplifiers,'' \emph{IEEE Transactions on
  Computer-Aided Design of Integrated Circuits and Systems}, vol.~22, no.~2,
  pp. 198--212, 2003.

\bibitem{li2019artificial}
Y.~Li, Y.~Wang, Y.~Li, R.~Zhou, and Z.~Lin, ``An artificial neural network
  assisted optimization system for analog design space exploration,''
  \emph{IEEE Transactions on Computer-Aided Design of Integrated Circuits and
  Systems}, vol.~39, no.~10, pp. 2640--2653, 2019.

\bibitem{liu2021parasitic}
M.~Liu, W.~J. Turner, G.~F. Kokai, B.~Khailany, D.~Z. Pan, and H.~Ren,
  ``Parasitic-aware analog circuit sizing with graph neural networks and
  bayesian optimization,'' in \emph{2021 Design, Automation \& Test in Europe
  Conference \& Exhibition (DATE)}.\hskip 1em plus 0.5em minus 0.4em\relax
  IEEE, 2021, pp. 1372--1377.

\bibitem{zhang2019circuit}
G.~Zhang, H.~He, and D.~Katabi, ``Circuit-gnn: Graph neural networks for
  distributed circuit design,'' in \emph{International Conference on Machine
  Learning}.\hskip 1em plus 0.5em minus 0.4em\relax PMLR, 2019, pp. 7364--7373.

\bibitem{quarles1994spice}
T.~Quarles, A.~Newton, D.~Pederson, and A.~Sangiovanni-Vincentelli, ``Spice 3
  version 3f5 user's manual,'' 1994.

\bibitem{jaegle2021perceiver}
A.~Jaegle, S.~Borgeaud, J.-B. Alayrac, C.~Doersch, C.~Ionescu, D.~Ding,
  S.~Koppula, D.~Zoran, A.~Brock, E.~Shelhamer \emph{et~al.}, ``Perceiver io: A
  general architecture for structured inputs \& outputs,'' \emph{arXiv preprint
  arXiv:2107.14795}, 2021.

\bibitem{jain2021representing}
P.~Jain, Z.~Wu, M.~Wright, A.~Mirhoseini, J.~E. Gonzalez, and I.~Stoica,
  ``Representing long-range context for graph neural networks with global
  attention,'' \emph{Advances in Neural Information Processing Systems},
  vol.~34, 2021.

\bibitem{li2020deepergcn}
G.~Li, C.~Xiong, A.~Thabet, and B.~Ghanem, ``Deepergcn: All you need to train
  deeper gcns,'' \emph{arXiv preprint arXiv:2006.07739}, 2020.

\bibitem{kipf2016semi}
T.~N. Kipf and M.~Welling, ``Semi-supervised classification with graph
  convolutional networks,'' \emph{arXiv preprint arXiv:1609.02907}, 2016.

\bibitem{velivckovic2017graph}
P.~Veli{\v{c}}kovi{\'c}, G.~Cucurull, A.~Casanova, A.~Romero, P.~Lio, and
  Y.~Bengio, ``Graph attention networks,'' \emph{arXiv preprint
  arXiv:1710.10903}, 2017.

\bibitem{hu2020open}
W.~Hu, M.~Fey, M.~Zitnik, Y.~Dong, H.~Ren, B.~Liu, M.~Catasta, and J.~Leskovec,
  ``Open graph benchmark: Datasets for machine learning on graphs,''
  \emph{arXiv preprint arXiv:2005.00687}, 2020.

\bibitem{fey2019fast}
M.~Fey and J.~E. Lenssen, ``Fast graph representation learning with pytorch
  geometric,'' \emph{arXiv preprint arXiv:1903.02428}, 2019.

\end{thebibliography}
\bibliographystyle{IEEEtran}

\end{document}